\documentclass[twocolumn]{article}

\usepackage{authblk}
\usepackage{times}
\usepackage{amsfonts}
\usepackage{graphicx}
\usepackage{csquotes}
\usepackage{siunitx}
\usepackage{color}
\usepackage{latexsym}
\usepackage{float}

\usepackage{tikz}
\usetikzlibrary{shapes,arrows, calc}
\tikzset{
	font={\fontsize{10pt}{12}\selectfont}}

\tikzstyle{decision} = [diamond, draw, fill=white!20,
text width=6em, text badly centered, node distance=3cm, inner sep=0pt]
\tikzstyle{block} = [rectangle, draw, fill=white!20,
text width=20em, text centered, rounded corners, minimum height=2em]
\tikzstyle{line} = [draw, -latex']
\tikzstyle{cloud} = [draw, ellipse,fill=white!20, text width=5em, text centered,
node distance=2.5cm, minimum height=2em]

\usepackage{listings}
\usepackage{color} 
\definecolor{mygreen}{RGB}{117, 199, 101}

\begin{document}

\title{A Compact Spectral Descriptor for Shape Deformations}

\author[1]{Skylar Sible}
\author[2]{Rodrigo Iza-Teran}
\author[2,3]{Jochen Garcke}
\author[4]{Nikola Aulig}
\author[4]{Patricia Wollstadt}

\affil[1]{The Ohio State University, Columbus, Ohio USA}
\affil[ ]{\textit{sible.3@buckeyemail.osu.edu}}
\affil[2]{Fraunhofer Center for ML and FhI. SCAI, Sankt Augustin, Germany}
\affil[ ]{\textit{\{rodrigo.iza-teran, jochen.garcke\}@scai.fraunhofer.de}}
\affil[3]{Institut f{\"u}r Numerische Simulation, Universit{\"a}t Bonn, Germany}
\affil[4]{Honda Research Institute Europe GmbH, Offenbach/Main, Germany}
\affil[ ]{\textit{\{patricia.wollstadt, nikola.aulig\}@honda-ri.de}}

\twocolumn[
  \begin{@twocolumnfalse}

	\maketitle

	\begin{abstract} 
		Modern product design in the engineering domain is increasingly driven by
		computational analysis including finite-element based simulation, computational
		optimization, and modern data analysis techniques such as machine learning. To
		apply these methods, suitable data representations for components under
		development as well as for related design criteria have to be found. While a
		component's geometry is typically represented by a polygon surface mesh, it is
		often not clear how to parametrize critical design properties in order to
		enable efficient computational analysis. In the present work, we propose a
		novel methodology to obtain a parameterization of a component's plastic
		deformation behavior under stress, which is an important design criterion in
		many application domains, for example, when optimizing the crash behavior in
		the automotive context. Existing parameterizations limit computational analysis
		to relatively simple deformations and typically require extensive input by an
		expert, making the design process time intensive and costly. Hence, we propose
		a way to derive a compact descriptor of deformation behavior that is based on
		spectral mesh processing and enables a low-dimensional representation of also
		complex deformations.We demonstrate the descriptor's ability to represent
		relevant deformation behavior by applying it in a nearest-neighbor search to
		identify similar simulation results in a filtering task. The proposed
		descriptor provides a novel approach to the parametrization of geometric
		deformation behavior and enables the use of state-of-the-art data analysis
		techniques such as machine learning to engineering tasks concerned with plastic
		deformation behavior.
	\end{abstract}
	\vspace{0.5cm}
  \end{@twocolumnfalse}
]

%
%
\section{INTRODUCTION}

Modern engineering product design relies heavily on computer-aided engineering
(CAE) methods such as finite-element based simulation or computational
optimization. With increasing computational capabilities and shorter design
cycles in many industries, CAE methods are applied more and more extensively.
As a result, a tremendous amount of data has become available, especially due
to many components being repeatedly re-designed or optimized. This data
provides potential to apply state-of-the-art machine learning techniques~\cite{Burrows2011,Garcke2017,Zhao2010} including deep learning approaches
\cite{Georgiou2018,Guo2018,Umetani2017} in order to increase the efficiency and
quality of the design process, but also to handle the amount of generated data
itself \cite{Burrows2011,Spruegel2018}. One obstacle are the unstructured and
finely detailed mesh representations typically used for design parts resulting in high-dimensional data vectors. Hence,
to enable the application of computational tools, designers have to find
suitable representations and parameterizations for both, components under
development as well as related design criteria
\cite{Graening2014,Spruegel2018}.

A critical design criterion in many application domains is the plastic
deformation of a component under stress. For example, in automotive design,
components are developed to exhibit a specific plastic deformation behavior
during a crash in order to protect vehicle occupants \cite{DuBois2004}. A
component's crash performance is improved if its plastic deformation avoids
intrusion into the passenger cabin or allows for high energy absorption
\cite{DuBois2004,Fang2017,Liu2014}. Hence, the ability to computationally
analyze or optimize for specific deformation behaviors is a powerful design
tool.

\begin{figure}[t]
	\centering
	\includegraphics[width=0.475\textwidth]{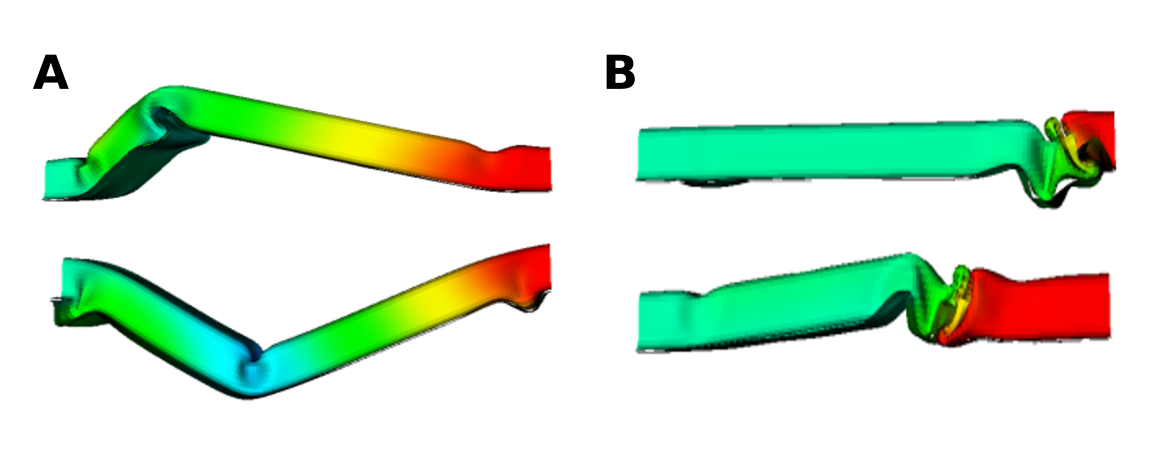}
	\caption{Modes of deformation typically encountered in automotive
	engineering \protect\cite{DuBois2004}:
	(A) Upward and downward bending deformation mode,
	(B) axial crushing or folding deformation mode.}
	\label{fig:deformation_modes}
\end{figure}

Current CAE-approaches typically study deformation behavior of a design part by
representing the part as a polygon surface mesh and quantifying the
displacement of a subset of selected nodes, for instance, during a crash
simulation \cite{Fang2017} (see also \cite{Redhe2004} for an example). Such a
parameterization is computationally efficient because it requires to only
monitor a small set of quantities of interest during the simulation---yet, it
is  limited in terms of the geometric complexity it is able to represent.
For example, a bending of a part may be easily quantified by the displacement
of a single node in a specified direction (Figure
\ref{fig:deformation_modes}A), whereas more complex deformations, such as an axial
crushing of the part, is not as easily quantified by the displacement of
pre-selected mesh nodes (Figure \ref{fig:deformation_modes}B). Hence, the
evaluation of deformation behavior, e.g., as a result of a crash simulation,
requires an expert to manually check the resulting deformation and to alter the
design if necessary. As a result, the development of a component requires many
iterations of simulation and visual inspection (e.g., \cite{Tang2016}), which
makes the design process costly.

In this paper we present a novel representation for geometric deformations
using a compact descriptor that allows for the representation of complex
deformation modes using methods from spectral mesh processing (e.g.,
\cite{Sorkine2005,Zhang2010}). Spectral mesh processing represents geometric
shapes as a set of coefficients with respect to an eigenbasis obtained from
spectral decomposition of a discrete mesh operator (see next section). We
propose to represent deformations by identifying the subset of spectral
coefficients, sufficient to describes a shape's plastic deformation with
respect to its original form, and present a workflow to select these relevant
spectral coefficients. In the following, we will first introduce spectral mesh
processing and its application in the engineering domain, before we formally
introduce our proposed method and demonstrate its application in a filtering
task, relevant to engineering contexts. We conclude with a discussion of the
results and directions for further research.

\section{SPECTRAL MESH PROCESSING IN THE ENGINEERING DOMAIN}

Spectral mesh processing is an approach to geometry processing that, similar to
Fourier analysis in the 1D-domain, is based on the eigendecomposition of a
suitable, discrete operator defined on the mesh (e.g., a discrete
Laplace-Beltrami operator \cite{Zhang2010}). The decomposition returns
eigenvectors that form an orthonormal basis into which any discrete function
defined on the mesh can be projected, yielding a set of spectral coefficients.
A common approach is to interpret the mesh vertices' Euclidean coordinates as
mesh functions and project them into the eigenbasis to obtain a spectral
representation of the mesh geometry. This transformation is invertible such
that the geometry's spatial representation can be reconstructed from its
spectral representation.

Representing mesh geometries in the spectral domain has interesting properties
and allows to address various geometric processing tasks more easily. For
example, obtained eigenvectors represent different geometric contributions to a
geometry, where eigenvectors associated with smaller eigenvalues describe
low-frequency functions on the mesh, while larger eigenvalues describe
high-frequency functions \cite{Sorkine2005} (where the ordering depends on the
operator used). This property is used in spectral mesh compression
\cite{Karni2000}, which efficiently stores geometries in terms of their first
$M$ low-frequency eigenvectors and corresponding spectral coefficients. This
approach assumes that most geometries can be approximated sufficiently by their
first $M$ low-frequency, spectral components. From this reduced set of spectral
components, the geometry's spatial representation can be reconstructed with
acceptable loss. Similar spectral approximations have been used to realize
shape matching and retrieval based on a subset of eigenvectors
\cite{Reuter2005,Reuter2006}. Selecting a subset of spectral components for
reconstruction using an orthogonal basis also relates to nonlinear approximation,
see e.g. DeVore \cite{devore}, and has also been used in image processing
\cite{Peyre}.

In the engineering domain, it has been shown that spectral mesh representations
using the Laplace-Beltrami operator are an adequate choice for representing
several deformations in a compact way
\cite{Garcke2017,IzaTeran2016Thesis,IzaTeran2018}. Here, an approach has been
proposed that uses the decomposition of one Laplace-Beltrami operator for a
series of deformations, which are assumed to be isometric to a base shape.
Spectral coefficients obtained by projecting several mesh deformations in $x$-,
$y$- and $z$-directions into a common eigenbasis are used as a dimensionality
reduction and visualization technique for large sets of deformed shapes
resulting from crash simulations. It has also been proposed that certain
eigenvectors can be interpreted as specific geometric operations on the shape,
e.g., a translation in the Euclidean space \cite{IzaTeran2018}. Furthermore,
the presented approach allows to handle arbitrary discrete functions defined on
the mesh, so that additional functional properties of a part, such as plastic
strain or stress, may be represented in the spectral domain.

Finally, note that in geometry processing the spectral representation of a
shape is often used to find pose invariant representations, i.e., to
distinguish a shape independent of pose changes by using the eigenfunctions
as shape features. On the other hand, in the studied engineering applications
part of the objective is to distinguish them in a pose dependent way, which is
achieved by projecting the deformations as functions into the (joint)
eigenbasis.

%
%
\section{PROPOSED METHOD}

Building on previous work in spectral mesh processing, the present paper
proposes to solve the problem of efficiently representing geometric
deformations by constructing a compact, spectral descriptor through the
targeted and adaptive selection of spectral components. In particular, we
propose to select the subset of only those components that are relevant for
representing application-specific deformations in the spatial domain,
and not just use the eigenvalue order as a fixed global criteria for the subset
selection. Relevant components can be identified by the relative magnitude of
their spectral coefficients, which indicates the importance of an eigenvector
for representing the geometry in the spatial domain. In other words, opposed to
previous applications such as mesh compression, we propose to select a subset of
spectral components not based on their corresponding geometric frequency, but
based on their contribution to the spatial representation irrespective of
frequency, as indicated by the magnitude of their spectral coefficients.

In particular, we propose a workflow to find a compact spectral descriptor for
deformations found in a set of geometries in three steps: a) compute a common
spectral basis that can be used for all simulations, b) identify deformation
modes in the set of simulations (e.g., through clustering in the spectral
domain), and  c) obtain spectral descriptors by identifying relevant spectral
components for each deformation mode.

The obtained spectral descriptor is significantly smaller than the full mesh
representation in the spatial domain, while preserving a high amount of
relevant geometric information, yielding an efficient representation of
geometric deformations for further computational analysis. Furthermore, we demonstrate
empirically that the resulting descriptor provides an abstract representation
of the deformation behavior by applying it in a nearest-neighbor search to
identify similar simulation results in a filtering task.

\subsection{Spectral mesh representation}


In order to describe the generation of the descriptor, we first provide a
formal definition of spectral mesh processing. Let $K=(G, P)$ be a
triangle mesh embedded in $\mathbb{R}^3$ with a graph $G=(V\!,\,E)$ describing
the connectivity of the mesh, where $V$ are mesh vertices with $|V|=N$ and $E
\subseteq V \times V$ the set of edges. The matrix $P \in \mathbb{R}^{N \times
3}$ describes the coordinates of mesh vertices in Euclidean space such that
each vertex has coordinates $\mathbf{p}_i = (x_i, y_i, z_i)$. We view $K$ as an
approximation of the Riemannian manifold $\mathcal{M}$, isometrically embedded
into $\mathbb{R}^3$. Furthermore, let $f: \mathcal{M} \rightarrow \mathbb{R}$
be a continuous function on $\mathcal{M}$. Evaluating this function at vertices
$V$ yields the discrete mesh function $f_K : K \rightarrow \mathbb{R}$.


We may now define a discrete, linear operator
on $\mathcal{M}$ (see \cite{Zhang2010} for a discussion of possible operators).
We here consider the Laplace-Beltrami operator, which is defined as the
divergence of the gradient in the intrinsic geometry of the shape and is a generalization of the Laplace operator to
Riemannian manifolds. The operator is invariant under isometric transformations,
i.e., transformations that preserve geodesic distances on the shape. In the
following, we thus assume that we operate on sets of shapes that are isometries
of each other in order to assume constant eigenbases between deformed shapes.
In practice, numerically $\epsilon$-isometries will be present, which will
result in only approximately the same eigenbasis. Spectral mesh decomposition
still can be used accordingly since under suitable conditions the Laplacians
only differ by a scaling factor in such a case~\cite{IzaTeran2018}.

The eigendecomposition of the operator returns eigenvectors $E=\left[\psi_1,
\psi_2, \ldots, \psi_N \right]$, ordered by the magnitude of the corresponding
eigenvalues, $\lambda_1<\lambda_2<\ldots<\lambda_N$, where each eigenvector
corresponds to a frequency component of the mesh function in increasing order
\cite{Sorkine2005}. 
Note that opposed to classical Fourier transform where basis functions are
fixed, the orthonormal basis obtained from the spectral decomposition of a mesh
operator depends on the mesh geometry and operator used. Often, only the set of
the first $M$ eigenvectors ordered by the magnitude of eigenvalues, $E_M$, is
used for further processing such that high-frequency components are discarded.

The normalized eigenvectors $E$ of a symmetric operator form an orthonormal
basis. We therefore may map any discrete mesh function $f_K$, given by a vector
$\mathbf{f}$, into this basis to obtain a representation, or encoding, in the spectral domain,

\begin{equation}
	\hat{\mathbf{f}} = E^\top\mathbf{f},
	\label{eq:spectrum_mesh_function_mat}
\end{equation}

\noindent where the columns of $E$ are eigenvectors $\{\psi_i\}^N_{i=1}$ and
$\hat{\mathbf{f}}$ contains corresponding spectral coefficients
$\{\alpha_i\}_{i=1}^N$. The spectral coefficients are thus obtained by
calculating $\{\alpha_i\}_{i=1}^N=\langle  \mathbf{f}, \psi_i\rangle$.
The inverse transform reconstructs, or decodes, the mesh function in the spatial domain,

\begin{equation}
	\mathbf{f} = E\,\hat{\mathbf{f}}.
	\label{eq:spectrum_mesh_function_mat_inverse}
\end{equation}

By considering Euclidean coordinates, $P = \left[ \mathbf{f}_x, \mathbf{f}_y,
\mathbf{f}_z \right] $, as mesh functions we project the mesh geometry into the
spectral domain,

\begin{equation}
	\hat{P} = E^\top P,
	\label{eq:spectrum_mesh_coords_mat}
\end{equation}

\noindent such that each row of $\hat{P}$, $\hat{\mathbf{p}}_i = \left[
\alpha^x_i, \alpha^y_i, \alpha^z_i \right], i=1,\ldots,N$, contains spectral coefficients
that can be used to express $x$-, $y$-, and $z$-coordinates as

\begin{equation}
	\mathbf{f}_x = \sum_{i=1}^N \alpha_i^x \psi_i ,\,\,
	\mathbf{f}_y = \sum_{i=1}^N \alpha_i^y \psi_i ,\,\,
	\mathbf{f}_z = \sum_{i=1}^N \alpha_i^z \psi_i.
	\label{eq:spectrum_cart_coords_reconstruction}
\end{equation}

An approximation of the mesh geometry is obtained by using the coefficients
corresponding to the first $M \ll N$ eigenfunctions, or suitably selected ones.

In a CAE-application, the spectral representation may now be used to project a
set of geometries, each represented as three-valued mesh functions, into the same spectral basis, allowing for a joint handling
of the data in a common space. This approach assumes that geometries are
available in a regular mesh format and that the mesh is the same, or is suitably
interpolated, for all geometries in the set. A single Laplace-Beltrami operator
is then computed for the set of deformations \cite{Garcke2017,IzaTeran2018}.
Observe that the operator is computed using geodesic distances along the surface of a
shape, where one assumes that the deformations do not modify this distance.
As a result,
the approach yields a common representation for deformations using the spectral
coefficients obtained by projecting three functions, each for mesh deformations in $x$-, $y$- and
$z$-directions, to the eigenvectors of only one shape.


We would like to explain some of the properties of the data representation by comparing to principal component analysis (PCA).
In that data-driven approach, a data matrix (e.g., comprising of the deformations as vectors) gets compressed into a small number components based on the variance of the data, where the largest variance will be contained in the first principal component.
As an example let us consider a series of deformations that puts two areas of a part nearby.
Here, the PCA will provide principal components that concentrate on those sections of the part and will reproduce behavior similar to this one.
Now, let us assume one did not include data for a deformation that affects only one area of the part.
The principal components will not be able to suitably represent such an unseen and strongly different deformation behavior since it was not trained for that.

On the other hand, take the basis obtained from the Laplace-Beltraim operator, where only the shape geometry is taken into account.
In this basis, the (new) deformation of the shape can be reproduced in the same fashion as the earlier ones.
In other words, whereas in the PCA higher variance, which can be interpreted as higher frequencies of the input data, is discarded, in the Laplace-Beltrami basis higher (geometric) frequencies of the underlying shape are discarded.
Note here also, it was shown using suitable assumptions that for functions bounded in the
$H^1$-Sobolev norm the $L_2$-approximation using the orthonormal basis obtained from the Laplace operator is optimal in a certain best basis sense~\cite{Brezis2017}; this result can be extended
to the Laplace-Beltrami operator and functions in the Sobolev space $H^{2,2}$
on the underlying manifold~\cite{Tesch2018}.

\subsection{Finding an efficient spectral descriptor for plastic deformations} 


In the spectral domain, individual eigenvectors, $\psi_i$, can be interpreted
as geometric contributions to the spatial representation of the shape
(geometric frequencies), where in general eigenvectors with low eigenvalues
represent more low-frequency contributions and those with high eigenvalues
high-frequency contributions (depending on the operator used). Furthermore,
eigenvectors may be associated with specific geometric transformations of the
shape, such that changes in the corresponding spectral coefficients can even
have a mathematical (e.g., rotation of a shape in the underlying space) or
physical interpretation (e.g., deformation in parts of the shape).

When projecting Euclidean coordinates, $P = \left[\mathbf{f}_x, \mathbf{f}_y,
\mathbf{f}_z \right] $, into the eigenbasis, the magnitude of resulting
spectral coefficients associated with each eigenvector,
$\{\alpha_i^x,\alpha_i^y,\alpha_i^z\}^N_{i=1}$, represents the relevance of
that eigenvector's geometric contribution to the deformed shape's  spatial
representation \cite{IzaTeran2018}. Based on this property, we propose to find
a compact spectral descriptor, $\mathbf{S}$, for a deformation, by selecting
only those coefficients, $\alpha^{(\cdot)}_j$, that have a high relative
magnitude compared to a suitable baseline. This is a further filtering in comparison to~\cite{IzaTeran2018} or mesh compression, where a fixed number of the spectral components is used, based on the order of the corresponding eigenvalues.
We assume that coefficients with
high values indicate that corresponding eigenvectors, $\psi_j$, represent
geometric information relevant for the description of the deformation in the
spatial domain.
Figure \ref{fig:workflow} shows an exemplary workflow for
finding $\mathbf{S}$, which is described in detail in the following.

\tikzstyle{block} = [rectangle, draw,
    text width=18em, text centered, minimum height=3em]
\tikzstyle{block2} = [rectangle, draw,
    text width=7em, text centered, minimum height=4em]
\tikzstyle{block3} = [rectangle, draw,
    text width=18em, text centered, minimum height=4em]
\tikzstyle{boxlabel} = [
    text width=8em]
\newcommand{\boxoffset}{0.4}
\newcommand{\scalefactor}{0.65}

\begin{figure}[ht!]
  \centering

	\begin{tikzpicture}[node distance = 1.6cm, auto, scale=\scalefactor, every node/.style={scale=\scalefactor}]
	\node [cloud] (start) {start};
  \node [block, below of=start] (selectgeom) {provide mesh representing desired deformation behavior};

  \node [block, below of=selectgeom, node distance=2cm] (operator) {calculate mesh operator};
  \node [block, below of=operator] (eigproblem) {calculate eigendecomposition of mesh operator};
  \node [block, below of=eigproblem] (spectralrep) {obtain spectral coefficients for mesh geometry};
  \node [boxlabel, left of=operator, node distance=5.5cm] at (0.1,-3.1) {\textcolor{red}{preprocessing}};

  \node [block, below of=spectralrep, node distance=2cm] (setthresh) {set initial $t$};
  \node [block, below of=setthresh] (identify) {obtain $\mathbf{S}$ by selecting coefficients $> t$};
  \node [block, below of=identify] (evaluate) {reconstruct spatial representation from $\mathbf{S}$};
  \node [decision, below of=evaluate] (decide) {spatial reconstruction sufficient?};
  \node [block2, left of=evaluate, node distance=5.5cm] (update) {adjust $t$};
  \node [boxlabel, left of=setthresh, node distance=5.5cm] at (0.1,-8.5) {\textcolor{blue}{descriptor construction}};

  \node [block, below of=decide, node distance=3cm] (return) {return $\mathbf{S}$};
  \node [block3, below of=return, node distance=2.2cm] (apply) {apply $\mathbf{S}$: calculate coefficients in $\mathbf{S}$ for new geometry};
  \node [boxlabel, left of=apply, node distance=5.5cm] at (0.1,-19.5) {\textcolor{mygreen}{application}};
	\node [cloud, below of=apply, node distance=2cm] (end) {end};
  \path [line] (start) -- (selectgeom);
  \path [line] (selectgeom) -- (operator);
	\path [line] (operator) -- (eigproblem);
	\path [line] (eigproblem) -- (spectralrep);
	\path [line] (spectralrep) -- (setthresh);
  \path [line] (setthresh) -- (identify);
	\path [line] (identify) -- (evaluate);
	\path [line] (evaluate) -- (decide);
	\path [line] (decide) -| node [near start] {no} (update);
	\path [line] (update) |- (identify);
  \path [line] (decide) -- node {yes}(return);
  \path [line] (return) -- (apply);
  \path [line] (apply) -- (end);
	\draw[red,     thick] ($(operator.north west)+(-4.1,\boxoffset)$)  rectangle ($(spectralrep.south east)+(0.44,-\boxoffset)$);
	\draw[blue,    thick] ($(setthresh.north west)+(-4.1,\boxoffset)$) rectangle ($(return.south east)+(0.44,-\boxoffset)$);
	\draw[mygreen, thick] ($(apply.north west)+(-4.1,\boxoffset)$)  rectangle ($(apply.south east)+(0.44,-\boxoffset)$);

	\end{tikzpicture}
  \caption{Exemplary workflow for finding and applying the proposed spectral shape
  descriptor: preprocessing (red box); construction of $\mathbf{S}$ by
  selection of spectral coefficients (blue box); application of selected
  descriptor (green box). See main text for detailed description.}
  \label{fig:workflow}
\end{figure}
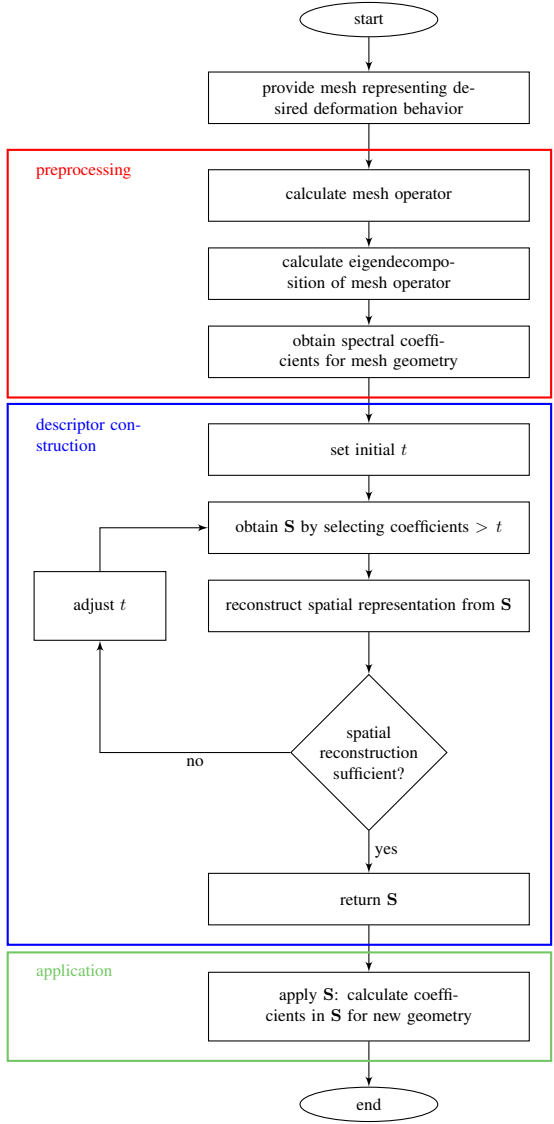

\begin{figure*}[ht]
	\centering
	\includegraphics[width=0.8\textwidth]{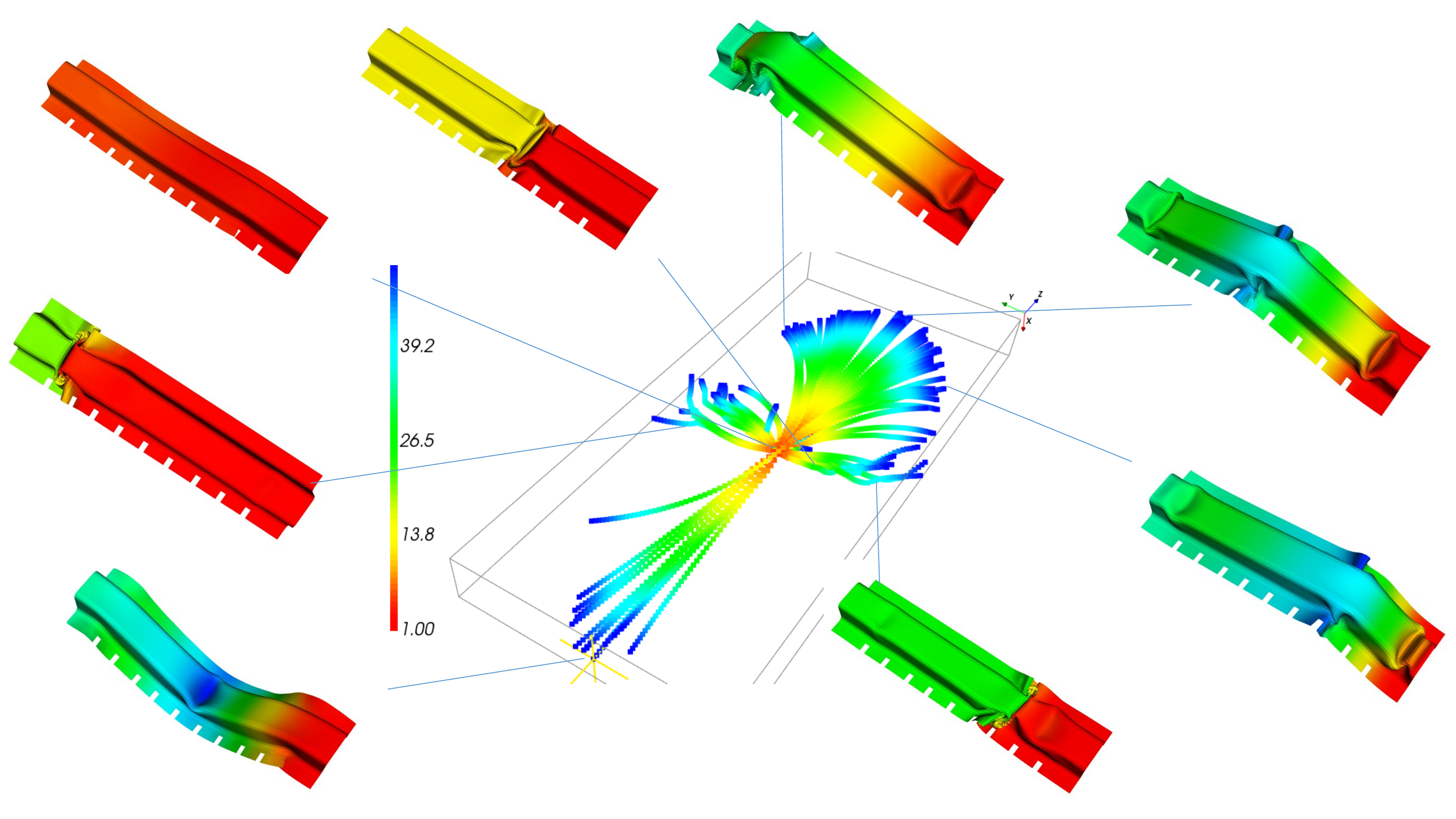}
	\caption{Visualization of the spectral coefficients for the first
	eigenvector of the $x$-, $y$-, $z$-coordinates for all simulation runs and
	all time steps (color coding indicates simulation time steps, where red
	denotes the first time step). Note how the spectral representation allows
	for an identification of the different deformation modes based on the
	distance of simulation runs in the spectral domain. Coefficients in the
	bottom left correspond to a downward bend, top right corresponds to upward
	bend, and left and right correspond to axial crushing along the length of
	the part. Color coding of parts indicates the difference to the undeformed
	baseline geometry, where red indicates smaller and blue larger distances.}
	\label{fig:spectral_basis_cluster_modes}
\end{figure*}

To find $\mathbf{S}$, we first have to provide a mesh representing the geometry
of a desired deformation behavior. For example, in the automotive context we
may wish to describe an axial folding of a beam during a frontal crash, because
this deformation mode leads to high energy absorption. Such a geometry may be
selected from a set of $k$ existing, different simulations. The desired
deformation may be identified either through manual selection by an expert
motivated by functional requirements, or through data-driven approaches in
either the spatial or spectral domain. For example, clustering of spectral
coefficients may be used to reveal main modes of deformations present in the
set of simulations (Figure \ref{fig:spectral_basis_cluster_modes}), e.g.,
bending or folding behavior (Figure \ref{fig:deformation_modes}). We may either
cluster the geometries based on their deformation in the final time step of a
simulation run, the transient data from the full simulation run over time, or
some other physical quantities such as plastic strain or stress on the surface
of the geometry.
In particular, a suitable visual representation of the coarse behavior of the
deformation can often be obtained by using the $x$-, $y$-, $z$-spectral
coefficients of the first eigenvector in eq.
(\ref{eq:spectrum_cart_coords_reconstruction}) \cite{Garcke2017,IzaTeran2018}.

Note that the application of the descriptor requires that the deformation, or
other quantities of interest, is sufficiently represented by the initial
surface mesh. In particular, the mesh resolution has to be fine enough to
represent the deformation. For application domains concerned with more
high-frequency deformations, this results in fine meshes with a high number of
nodes. Here, the applicability of the method may be limited by the practical
run time of the proposed method, which depends on the number of nodes in the
mesh (see also the asymptotic run time of the proposed approach, discussed in
section \textit{Computational complexity of the proposed workflow}).

Once a mesh representing the target deformation is identified, we calculate its
spectral representation according to eq. (\ref{eq:spectrum_mesh_coords_mat})
(Figure \ref{fig:workflow}, red box) to obtain coefficients
$\hat{\mathbf{p}}_i$. Now, several approaches to defining $\mathbf{S}$ by
selecting relevant coefficients are conceivable. In many applications, e.g.,
computer graphics, typically the first $M$ eigenfunctions are used. In contrast
to this approach, we propose to identify relevant eigenfunctions based on the
magnitude of their \textit{coefficients}. For a single shape exhibiting the
targeted deformation, we may thus select all coefficients exceeding a threshold
$t$,

\begin{equation}
	\mathbf{S} = \left\{\alpha_j^x,\alpha_j^y,\alpha_j^z | \alpha_j^x > t \vee \alpha^y_j > t \vee \alpha^z_j > t\right\},
	\label{eq:descriptor_single}
\end{equation}

\noindent where $t$ may be set based on a statistical criterion, e.g., relative
to the mean of coefficients, to identify those coefficients that indicate a
high relevance for representing the shape.

Alternatively, the shape describing the desired deformation may be contrasted
against the undeformed baseline shape, such that relevant coefficients can be
identified by the largest differences in coefficients between baseline and
deformed shape (Figure \ref{fig:data_generation}B,C). Here again the setting of
a threshold $t$ for selecting the highest differences is required.

Both procedures identify components that contribute to the spatial
representation of the targeted deformation while ignoring less relevant
geometric components. Hence, we obtain a sparse description of the geometric
information relevant to characterize a specific deformation in the spatial
domain.



For both approaches, $t$ may be either set based on a statistical criterion,
but may also be found through an iterative process that alternates between
lowering or increasing $t$ and evaluating the reconstruction quality in the
spatial domain (Figure \ref{fig:workflow}, blue box). Reconstruction quality
may either be judged through visual inspection by an expert or through
calculation of an error metric between the original and reconstructed shape.
Adjusting $t$ controls for the size of the descriptor, $|\mathbf{S}| = M$, and
thus allows for a trade-off between compactness and the level of geometric
detail captured by the descriptor. Note that recovering the original mesh
representation requires \textit{all} spectral coefficients (according to eq.
\ref{eq:spectrum_mesh_function_mat_inverse}), hence---even though  $\mathbf{S}$
is sufficient to represent the geometric deformation of a part---it is not
possible to recover the original mesh representation from $\mathbf{S}$ alone.
Nevertheless, an approximate reconstruction is still possible.

In our experiment, we demonstrate that, for a part typically encountered in
engineering contexts, setting $t$ based on a statistical criterion resulted in
a small descriptor size, $M \ll N$, where $N$ is the number of nodes in a
triangular surface mesh. This descriptor was able to robustly filter simulation
results based on their deformation mode despite the considerable reduction in
description length, indicating that a high amount of application-relevant
geometric detail was retained.

\subsection{Application scenarios}

Once the descriptor $\mathbf{S}$ is constructed, it can be applied in further
computational tasks (Figure \ref{fig:workflow}, green box). For example, the
descriptor may be used as a feature in a machine learning task, encoding the
plastic deformation behavior of a part. A possible application is
meta-modelling, which aims at replacing costly simulations or optimization runs
by a cheaper evaluation of statistical models. The descriptor can be used in
case the design process investigates the relationship between a geometry and an
objective function of interest. Learning such a meta-model requires the
efficient parametrizations of the properties of interest, such as deformation
behavior. Further tasks may include optimization of the part with respect to
some property, e.g., material thickness, while the shape descriptor is used as
a constraint or part of the objective function to ensure the desired
deformation behavior. Here, the descriptor can be used to describe the targeted
deformation behavior and to describe the deformation behavior of parts in
intermediate steps of the optimization. As soon as the desired and actual
deformation behavior diverge, measurable by an increasing distance between the
descriptors, the optimization can be stopped automatically. Alternatively, the
distance may be used as part of the objective function, e.g., when using
evolutionary optimization techniques.

A further task, common in the engineering design process, is the filtering and
verification of simulation results based on geometric properties. Often, large
numbers of simulation runs are performed in order to investigate the impact of
variations in design parameters on the component under development. This
results in large amounts of simulation data that typically require
time-intensive visual inspection by an expert in order to verify the success of
individual simulations \cite{Zhao2010,Burrows2011}. Our descriptor can be used
both, to automatically verify the outcome of a simulation with respect to the
desired deformation behavior, and to filter simulation results based on the
deformation behavior (see section \textit{Experiments}). In particular, the
proposed workflow may again be used to find the spectral descriptor for the
targeted deformation behavior and the actual deformation behavior of the part
being optimized. To verify a result, a threshold on the acceptable distance
between both descriptors can be set, allowing for the verification of large
sets of simulation results without the need for manual inspection by an expert.

Note that since the Laplace-Beltrami operator is only invariant under isometric
transformations, the presented approach is limited to scenarios where all
shapes involved in finding and using the descriptor represent the same baseline
geometry in different states of (isometric) deformation. This is however the
case for many engineering applications, where properties of the structure are
determined by variables that can be varied independently of the mesh geometry.
Two examples are thickness of shell finite elements or material properties of
finite elements such as failure criteria. Another application scenario is the
variation of loading or boundary conditions whose variation can lead to
different deformation behaviors while the base geometry of the component of
interest stays the same. In these scenarios, our method has relevance for
robustness or optimization studies, where certain deformation modes are desired
as a main objective. Additionally, small changes to the geometry can be handled
by interpolating the mesh to a joint reference mesh.

\subsection{Computational complexity of the proposed workflow}

In terms of computational efficiency, the most costly operations in
constructing the descriptor are the preprocessing phase (Figure
\ref{fig:workflow}, red box) comprising the calculation of the mesh operator,
its eigendecomposition, and the projection of Euclidean coordinates into the
eigenbases, eq. (\ref{eq:spectrum_mesh_coords_mat}). The asymptotic time
complexity of the preprocessing is dominated by the matrix multiplications
performed as part of the operator definition and its eigendecomposition, which
is cubic in the number of mesh vertices, $\mathcal{O}(N^3)$. This operation is
done only once and can be performed offline with respect to the application of
the descriptor in a subsequent task (Figure \ref{fig:workflow}, green box). The
time complexity is cubic if a naive algorithm is used, but also algorithms with
subcubic runtimes are available (e.g., \cite{Cormen2009}). Furthermore, since
the discrete Laplace-Beltrami matrix is inherently data sparse, fast
computations of the first, say, \num{1000} eigenvectors using numerical
approaches exploiting this data sparseness seem possible, which is part of
future work. In any case, this one-time pre-processing step is small in
comparison to the runtime of a single numerical simulation performed to
generate one datum.

The application of the descriptor to new geometries in the online phase (Figure
\ref{fig:workflow}, green box) requires the projection of the new geometry's
Euclidean coordinates, $P$, into the bases indicated by the descriptor,
$E_{\mathbf{S}}$. The projection consists of a matrix multiplication,
$E_{\mathbf{S}}^TP$, eq. (\ref{eq:spectrum_mesh_coords_mat}), which has
asymptotic time complexity $\mathcal{O}(3MN)$, with $M \ll N$.


\section{EXPERIMENTS}

\subsection{Data generation}

To demonstrate the effectiveness of our method, we used the proposed descriptor
to represent the deformation behavior of a hat section beam in a crash
simulation and filter simulation results based on deformation mode.

\begin{figure}[ht]
	\centering
	\includegraphics[scale=0.8]{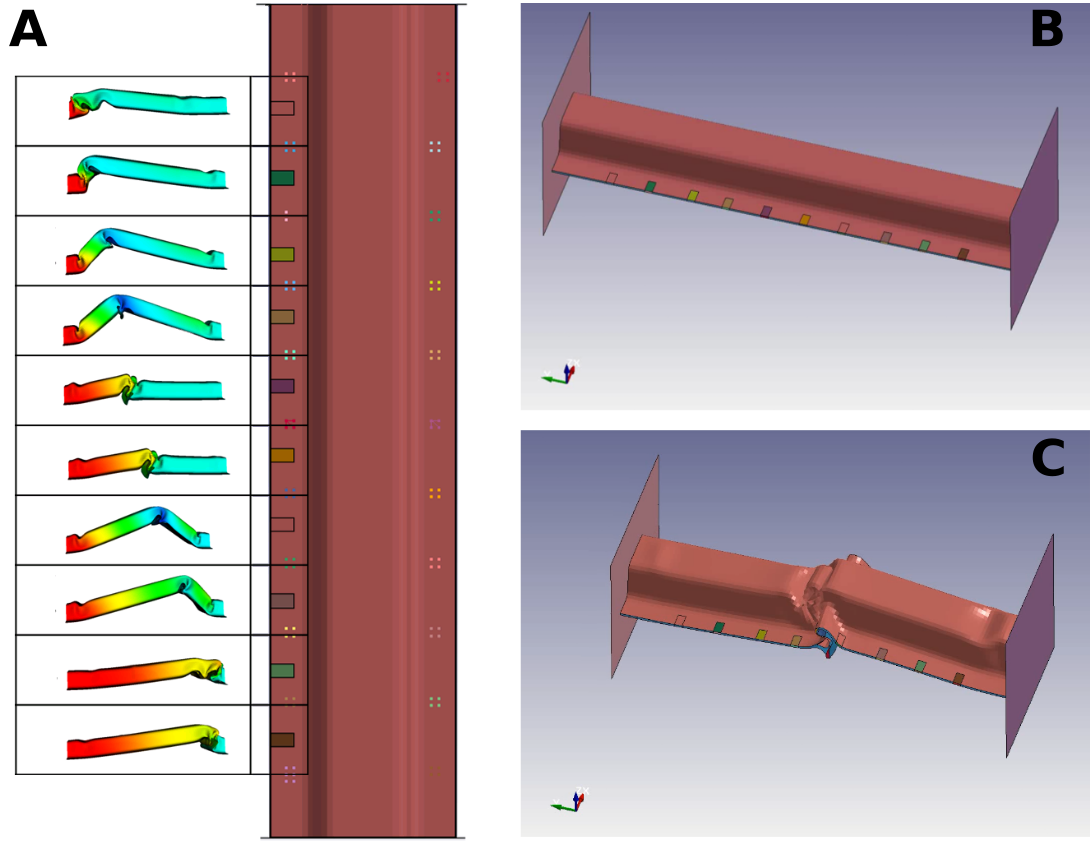}
	\caption{
	(A) Hat section beam used for data generation. Colored markers
    indicate notches introduced at various locations along the length of the
    shape, which lead to various different deformation behaviors shown to the
    left (notches denote areas of material thickness of
    \SI{0.0001}{\milli\meter}).
	(B) Simulation setup with undeformed baseline shape.
	(C) Exemplary final time step of simulation run showing deformed shape.}
	\label{fig:data_generation}
\end{figure}

A hat section beam is a structural element common in engineering domains such
as civil or automotive engineering (Figure \ref{fig:data_generation}A). It
consists of a top-section with a hat-shaped cross-section that is joined
together with a base plate.

We simulated an axial crush of the beam
using LS-DYNA mpp R7.1.1 (Figure \ref{fig:data_generation}B,C), double
precision, while inducing various deformation modes. On the resulting data, we
used the proposed descriptor to filter results based on a desired deformation
behavior.

Variation in the deformation behavior was introduced by adding
\enquote{notches} in the flange of the hat section at defined locations along
the length of the part (Figure \ref{fig:data_generation}A) and varying the
material thickness of the part between \SI{1}{\milli\meter} and
\SI{10}{\milli\meter}. A notch was simulated by setting the material thickness
to \SI{0.0001}{\milli\meter} in order to simulate a removal of the material at
that point. The setup resulted in a simulation bundle consisting of $k=100$
simulations with a large variety of deformation modes, each with several saved
time steps.


\subsection{Calculation of mesh spectral coefficients and selection of shape descriptor}

For further analysis, we considered the set of deformed shapes from the final
time step of all simulation runs. The spectral representation according to
eq. (\ref{eq:spectrum_mesh_coords_mat}) was computed, where we used the first \num{500} eigenvectors
with smallest eigenvalues.
We selected geometries representative of one
of three deformation modes, namely \emph{upward bend}, \emph{downward bend},
and \emph{axial crush} (Figure \ref{fig:deformation_modes}A, B), identified
through clustering of the first \num{500} coefficients in the spectral domain.
Alternatively, Figure \ref{fig:spectral_basis_cluster_modes} shows the spectral coefficients of
$x$-, $y$-, $z$-coordinates corresponding to the first eigenvector for all
simulation runs and time steps, from this visualization the different modes could also be selected.

\begin{figure}[H]
	\centering
	\includegraphics[width=0.35\textwidth]{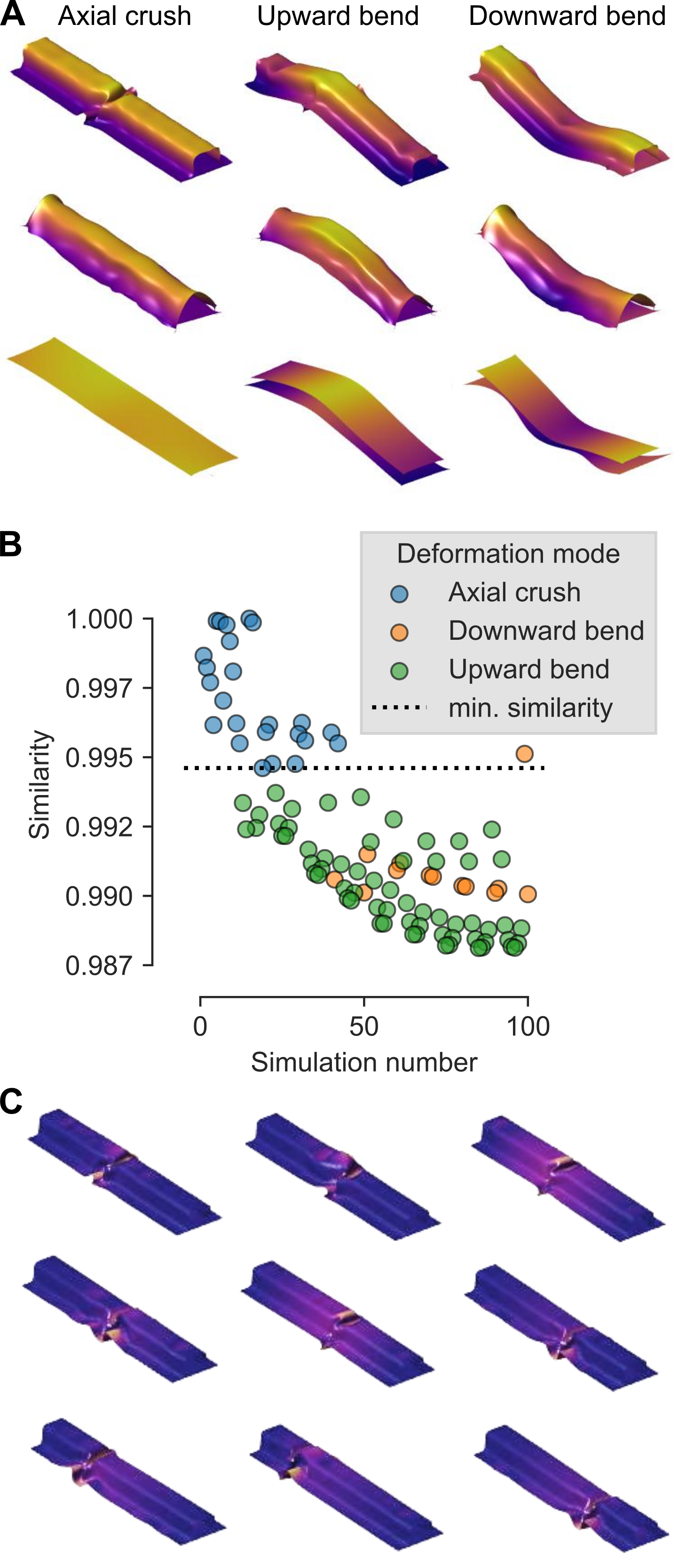}
	\caption{(A) Mesh reconstruction for three deformation modes (color scale
    indicates displacement in $z$-direction): top row shows mesh
    reconstructions from the first 500 eigenvectors ordered by magnitude of
    eigenvalues; middle row shows reconstruction from the proposed descriptor
    with sizes $M=14$ for axial crush, $M=17$ for upward bend, $M=16$ for
    downward bend, bottom row shows reconstruction from the first $M=14$,
    $M=17$, and $M=16$ eigenvectors ordered by magnitude of their eigenvalues.
    (B) Cosine similarity between axial crush descriptor and all simulation
    runs. (C) First nine most similar simulation results for axial crush
    spectral descriptor (color scale indicates displacement in $z$-direction).
	}
	\label{fig:experiments}
\end{figure}

For each deformation mode we obtained the spectral descriptor according to eq.
(\ref{eq:descriptor_single}) by setting $t$ to one standard deviation above the
mean over all coefficients for a representative shape. This approach resulted
in descriptor sizes of $M=14$ for axial crush, $M=17$ for upward bend, $M=16$
for downward bend, compared to an original mesh size, $N$, of around \num{9000}
nodes. The quality of the descriptor was validated through visual inspection of
the spatial reconstruction (Figure \ref{fig:experiments}A). Note that to allow
for proper reconstruction, for each selected coefficient, $\alpha^{(\cdot)}_j
\in \mathbf{S}$, also the spectral coefficients corresponding to the remaining
two 3D coordinates were added if not already contained in the descriptor.
Furthermore, eigenvectors $\psi_1$ and $\psi_2$ were added to the
reconstruction.


We compared the spatial reconstruction of the meshes from our descriptor of
length $M$ to the reconstruction from the first $M$ low-frequency components
(eigenvectors with smallest eigenvalues), i.e. $N=M$ in eq.
(\ref{eq:spectrum_cart_coords_reconstruction}) (Figure \ref{fig:experiments}A,
bottom row). The latter is a common approach in dimensionality reduction or
compression applications \cite{Karni2000,Garcke2017}. The comparison shows that
our descriptor obtained a better reconstruction quality than the reconstruction
from low-frequency components alone. In particular, the high-frequency
components included in the descriptor captured also finer detail on the mesh,
for example, the edge of the hat section in the center of the bent parts. An
approximate decoding from the sparse encoding of size $M$ is therefore
possible, where the quality is good enough to reconstruct the main behavior,
but not the details, e.g., the location of the axial crush is not preserved.

\subsection{Application of spectral descriptor for filtering and shape retrieval}

We used the spectral descriptors to filter the set of simulation results for
geometries exhibiting a specific crash behavior. As described in section
\textit{Application scenario}, filtering large-scale simulation runs based on
geometric properties is a common application scenario in the engineering design
process.

To filter simulation runs using the proposed descriptor, $\mathbf{S}$, we first
obtained spectral coefficients for all shapes through projection into the
spectral domain; we then calculated the similarity between these coefficients
and coefficients in $\mathbf{S}$ using the cosine similarity. Figure
\ref{fig:experiments}B shows the similarity between the descriptor of the axial
crush deformation mode and all simulation results. Our approach correctly
identified all simulation results exhibiting an axial crushing as most similar
to the spectral descriptor, with the exception of one geometry showing an
upward bend. Figure \ref{fig:experiments}C shows the nine simulation results
most similar to the descriptor, which all show an axial crushing of the beam.
Note that the approach was able to identify simulation results exhibiting an
axial crushing irrespective of the exact location of the axial folding along
the part.

In summary, the descriptor provided an abstract description of the
deformation behavior that did not require the specification of an exact
deformation, e.g., in terms of the displacement of individual nodes. The
descriptor successfully represented application-relevant geometric information
through the targeted selection of coefficients while being of much smaller size
than the full mesh representation in the spatial domain.

\section{CONCLUSION AND FUTURE WORK}

We proposed a novel approach for the efficient representation of geometric
deformations using a spectral descriptor comprising components selected in a
targeted and adaptive fashion. The selection procedure ensures that only
geometric components relevant for specifying the targeted deformation are
included, which makes the descriptor much smaller in size than a full geometric
representation of the deformation, e.g., by a surface mesh, and also smaller
and more focused than using the first few hundred spectral components as
in~\cite{IzaTeran2018}. Despite its compactness, the descriptor was able to
capture a high amount of application-relevant geometric information and
provided the necessary descriptive power to distinguish between deformation
modes in a filtering task. The trade-off between size and represented geometric
detail makes the descriptor a promising tool for the parametrization of also
complex geometric deformations in various computational tasks.

Future work may explore the applicability of the descriptor, for example, in
tasks such as structural optimization with plastic deformation as design
criterion. Here, the proposed descriptor offers a powerful design tool and may
improve on current approaches that require extensive manual intervention and
evaluation by an expert. Furthermore, the descriptor may be used in
post-processing of simulation data by identifying or filtering results based on
geometric deformation. Especially in large-scale data sets, such an approach
allows to automatically verify the success of a simulation instead of requiring
the visual inspection of large sets of simulation results by an expert.
Lastly, the descriptor may be used as a feature in machine
learning tasks on large-scale engineering data sets, representing a central
design property in many application domains.

\section*{Acknowledgement}
The authors thank Emily Nutwell of the OSU SIMCenter for support in generating
the hat section model.

\bibliographystyle{apalike}
\bibliography{deformation_modes}
\end{document}